\documentclass[letterpaper,12pt]{article}

\usepackage{arxiv}

\usepackage[utf8]{inputenc} 
\usepackage[T1]{fontenc}    
\usepackage{hyperref}       
\usepackage{url}            
\usepackage{booktabs}       
\usepackage{amsfonts}       
\usepackage{nicefrac}       
\usepackage{microtype}      
\usepackage{lipsum}
\usepackage{cite} 
\usepackage{amsmath}  
\usepackage{mathtools}
\usepackage{mathrsfs}
\usepackage{subfigure}
\usepackage[final]{graphicx}
\usepackage{algorithm,algpseudocode}
\usepackage{bbm}

\title{Variational Bayes: A report on\\ approaches and applications}
\newcommand*\samethanks[1][\value{footnote}]{\footnotemark[#1]}

\author{
  Manikanta Srikar Yellapragada\thanks{Equal contribution} \\
  New York University\\
  \texttt{msy290@nyu.edu} \\
   \And
 Chandra Prakash Konkimalla\samethanks  \\
  New York University\\
  \texttt{cpk290@nyu.edu} \\
}

\begin{document}
\maketitle

\begin{abstract}

Deep neural networks have achieved impressive results on a wide variety of tasks. However, quantifying uncertainty in the network's output is a challenging task. Bayesian models offer a mathematical framework to reason about model uncertainty. Variational methods have been used for approximating intractable integrals that arise in Bayesian inference for neural networks. In this report, we review the major variational inference concepts pertinent to Bayesian neural networks and compare various approximation methods used in literature. We also talk about the applications of variational bayes in Reinforcement learning and continual learning. 

\end{abstract}

\section{Introduction}
The effectiveness of deep neural networks has been rigorously demonstrated in the domains of computer vision, speech recognition and natural language processing, where they have achieved state of the art performance. However, they are prone to overfitting; spurious patterns are found that happen to fit well in training data, but don't work for new data. When used for supervised learning or reinforcement learning, they tend to make overly confident decisions about the output class and don't estimate uncertainty of the prediction. \\ \\
On the other hand, Bayesian Neural Networks  can learn a distribution over weights and can estimate uncertainty associated with the outputs. Markov Chain Monte Carlo (MCMC)  is a class of approximation methods with asymptotic guarantees, but are slow since it involves repeated sampling. An alternative to MCMC is variational inference, in which the posterior distribution is approximated by a learned variational distribution of weights, with learnable parameters. This is done by minimizing the Kullback-Leibler divergence between the posterior and the approximating distribution. This methodology and its applications in reinforcement learning and continual learning will be reviewed in this report. \\ \\
The rest of the report is structured as follows: Section 2 highlights the key theoretical aspects involved in Variational methods. Section 3 briefly describes the methods used for approximating Bayesian inference in NNs such as Practical Variational Inference for Neural Networks \cite{graves2011practical}, Auto-encoding variational bayes \cite{kingma2013auto}, Bayes by Backprop \cite{blundell2015weight}, Bayesian Hypernetworks \cite{krueger2017bayesian} and Multiplicative normalizing flows \cite{louizos2017multiplicative}. Section 4 talks about the  applications of weight uncertainty in exploration in Reinforcement learning such as Noisy networks \cite{fortunato2017noisy}, Deep Exploration via Bootstrapped DQN \cite{osband2016deep}, UCB Exploration via Q-Ensembles \cite{chen2017ucb} and in contintual learning such as  Variational continual learning \cite{nguyen2017variational}. 
\section{Formal background}
This section highlights the key definitions and theoretical aspects involved in variational bayesian methods and reinforcement learning. 

\subsection{Bayesian inference}
Bayesian inference is a method of statistical inference in which Bayes' theorem is used to update the probability of a hypothesis upon observing data. \\\\
\textbf{Definition 1 (Bayes' Theorem):}The Bayes' theorem is stated as the following equation: 
\begin{align*}
    P(Z | X) = \frac{P(X | Z) P(Z)}{P(X)} 
\end{align*}
\begin{itemize}
    \item $P(Z)$ is the prior probability, which is the probability of the hypothesis $Z$ being true before Bayes' theorem is applied. It is generally easy to compute as it's just a prior distribution that can be defined as a tractable function. In some sense, the prior contains all the knowledge we know thus far.
    \item $P(X | Z)$ is the likelihood function, which is the conditional probability of evidence $X$ given a hypothesis $Z$. 
\item $P(Z | X)$  is the posterior, which is what we actually want to compute or learn.
    \item $P(X)$ is the marginal likelihood term which denotes the probability of evidence or data. It can be calculated using the law of total probability as:
\begin{align*}
    P(X) &= P(X|Z_1) P(Z_1) + P(X|Z_2) P(Z_2) + ... P(X|Z_n) P(Z_n) \\
    &=  \sum_{Z_i \in Z} P(X|Z_i) P(Z_i)
\end{align*}
When there are an infinite number of outcomes (continuous random variables) , it is necessary to integrate over all outcomes to calculate $P(X)$ as:
\begin{align*}
    P(x) &= \int_{Z} P(x|z) P(z) dz
\end{align*}
\end{itemize}
\subsection{Entropy and KL divergence}

\textbf{Definition 2 (Entropy):}The entropy for a probability distribution defined as the amount of information present in it.
\begin{align*}
    \mathcal{H}(p) = - \sum_{i = 1}^N p(x_i) \log p(x_i)
\end{align*}
Entropy can be looked at as the minimum number of bits needed to encode an event drawn from a probability distribution. For example, for an eight-sided die where each outcome is equally probable, $\sum_1^8 \ln \frac{1}{8} = 3 $ bits are required to encode the roll. Kullback-Leibler (KL) Divergence is a slight modification to the formula for entropy. Instead of having just a probability distribution P, an approximating distribution Q is added.  \\\\
\textbf{Definition 3 (KL divergence):} The KL divergence is a measure of how one probability distribution is different from a second probability distribution.
\begin{align*}
    D_{KL}(P||Q) = \sum_{x \in X} P(x) \log \frac{P(x)}{Q(x)}
\end{align*}
For continuous random variables, KL divergence is given by the integral:
\begin{align*}
    D_{KL}(p||q) = \int_{- \infty}^\infty p(x) \log \frac{p(x)}{q(x)} dx
\end{align*}
\begin{itemize}
    \item $KL(p||q) \geq 0$ for all $p,q$
    \item $KL(p||q) = 0$ only if $p = q$
    \item $KL(p||q) \neq KL(q||p)$
\end{itemize}
\subsection{The problem of approximate inference}
The integral $P(X) = \int_{Z} P(X|Z) P(Z) dZ$ is typically intractable. There are two commonly used methods to solve Bayesian inference problems: MCMC and variational inference. MCMC is quite slow since it involves repeated sampling. Variational inference however is much faster as it can be stated as an optimization problem.  \\\\
Variational inference seeks to approximate the true posterior $P(Z | X)$ by introducing a new distribution $q(Z)$ that is as close as possible to the true posterior. The approximate distribution can have their own variational parameters : $q(Z|\theta)$ and we try to find the set of parameters that bring q closer to the true posterior. To measure the closeness of $q(Z)$ and $p(Z|X)$, KL divergence is used. The KL divergence for variational inference is:
\begin{align}
    KL (q(z) || p(z|x)) =& \int_Z q(z) \log \frac{q(z)}{p(z|x)} dz \\
    =& - \int_Z q(z) \log \frac{p(z|x)}{q(z)} dz \\
    =& - \Big(\int_Z q(z) \log \frac{p(x,z)}{q(z)} dz  - \int_Z q(z) \log p(x) dz \Big ) \\
    =& -\int_Z q(z) \log \frac{p(x,z)}{q(z)} dz + \log p(x) \int_Z q(z) dz \\
    =& -\mathcal{L}(x) + \log p(x)
\end{align}
where $\mathcal{L}(x)$ is known as the variational lower bound  or Evidence Lower Bound (ELBO). Equation (5) is obtained because $\int_Z q(z) dz = 1$ If we further decompose ELBO, we get:
\begin{align}
    \mathcal{L}(x) =& \int_Z q(z) \log \frac{p(x,z)}{q(z)} dz \\
    =& \int_Z q(z) \log p(x|z) - q(z) \log \frac{q(z)}{p(z)} dz \\
    =& E_q [\log p(z|x)] - KL(q(z) || p(z)) \\
    =& \int_z q(z) \log p(x,z) - q(z) \log q(z) dz \\
    =& E_q [\log p(x,z)] + \mathcal{H}(q)
\end{align}
where $\mathcal{H}(q)$ is the entropy of q. The first term in Equation (10) represents an energy, which encourages $q(z)$ to focus the probability mass where the model has high likelihood $p(x,z)$. The second term, entropy, encourages  $q(z)$ to diffuse across the space and avoid concentrating to one location.

\subsection{Minimum Description Loss}
As we know that in Variational Inference we try to minimize the Variational Free Energy($\mathcal{F}$) which is defined by 
\begin{align}
    \mathcal{F} = -\bigg\langle ln\bigg[\dfrac{P(x,y/w)P(w/\alpha)}{q(w/\beta)}\bigg] \bigg\rangle_{w \sim q(\beta)}
\end{align}
Here $P(w/\alpha)$ are prior probability of parameters. We can rearrange the terms of Equation (11) to get minimum description length loss function which is

\begin{align}
    \mathcal{F} &= \big\langle L^N(x,y,w) \big\rangle_{w \sim q(\beta)} +
         KL(q(\beta)||P(\alpha))\\
    where, L^N(x,y,w) &= - \sum_{(x,y)\in D} lnP(y/x,w)
\end{align}

We can rewrite Equation (12) as 
\begin{align}
    L(\alpha,\beta,D) &= L^E(\beta,D) + L^C(\alpha,\beta)\\
    where, L^E(\beta,D) &= L^N(x,y,w) ;  and \\
    L^C(\alpha,\beta) &= KL(q(\beta)||P(\alpha)) 
\end{align}
Here $D$ is the dataset or tuple$(x,y)$,$q(\beta)$ is the posterior, $\alpha$ and $\beta$ are prior parameters and posterior parameters respectively.

\subsection{Normalizing flows}

A normalizing flow consists of the transformation of one probability distribution into another probability distribution through the application of a series of invertible mappings. Possible applications of a normalizing flow include generative models, flexible variational inference, and density estimation, with issues such as scalability of each of these applications depending on the specifics of the invertible mappings involved. \\\\
\textbf{Change of variable:}
Consider an invertible smooth mapping $f: \mathbb{R}^d \to \mathbb{R}^d$ with inverse $f^{-1} = g$. If we use this mapping to transform a random variable $z \in \mathbb{R}^d$ with distribution $q(z)$, the resulting random variable $y$ would have a probability distribution: 
\begin{align*}
    q(y) = q(z) \Big| det \frac{\partial f}{\partial z} \Big| ^{-1}
\end{align*}
Introduced in \cite{rezende2015variational}, if a series of mappings $f_k, k \in 1 ,2,.., K $ are applied, the resulting probability density $q_K(y)$ would be 
\begin{align*}
    z_K &= f_K \circ ... f_2 \circ f_1 (z_0) \\
    \ln q_K(z_K) &= \ln q_0 (z_0) - \sum_{k=1}^K \ln  \Big| det \frac{\partial f_k}{\partial z_{k-1}} \Big| 
\end{align*}
\subsection{Reinforcement learning}
The idea behind Reinforcement Learning is that an agent will learn from the environment (everything outside the agent) by interacting with it and receiving rewards for performing actions. It is distinguished from other forms of learning as its goal is only to maximise the reward signal without relying upon some predefined labelled dataset. The agent tries to figure out the the best actions to take or the optimal way to behave in the environment in order to carry out his task in the best possible way. \\\\
\textbf{Definition 4 (Markov Decision Process):} A Markov decision process is a tuple $(S,A,\mathcal{P},R,\gamma)$ such that 
\begin{align*}
    \mathcal{P}(s',r|s,a) = Pr\{S_{t+1} = s', R_t = r|S_{t} = s, A_{t} = a\}
\end{align*}
where $S_t \in S$ (state space), $A_t \in A$ (action space) and $P$ is the state transition probability function. R is the reward function, and $\gamma$ is the discount factor $\gamma \in [0,1]$.

\begin{figure}[ht]
    \centering
    \includegraphics[width=9cm]{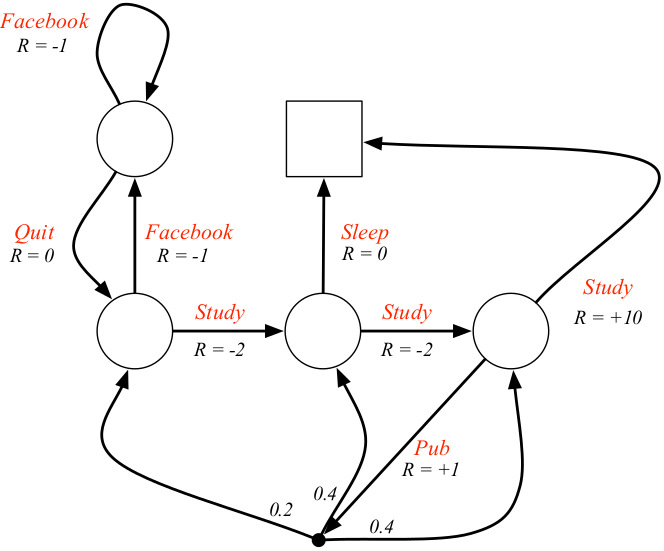}
    \caption{Example of a simple MDP from David Silver's lecture}
    \label{fig:mdp}
\end{figure}
\textbf{Definition 5 (Policy):} A policy $\pi$ is a distribution over actions given states. A policy fully defines the behavior of an agent.
\begin{align*}
    \pi (a|s) = \mathcal{P}(A_t = a| S_t = s)
\end{align*}
\textbf{Definition 6 (State Value Function):} The State Value function $V^\pi(s)$ of an MDP is the expected return starting from state S, and then following policy $\pi$.
\begin{align*}
   V^\pi(s) &= E_\pi [G_t | S_t = s] \\
   & = E_\pi [ \sum^\infty_{k=0} \gamma^k R_{t+k+1} | S_t = s]
\end{align*} 
\textbf{Definition 7 (Action Value Function):} The action-value function $Q^\pi (s,a)$ is the expected return starting from state s, taking action a, and following policy $\pi$. It tells us how good is it to take a particular action from a particular state.
\begin{align*}
   Q^\pi(s,a) &= E_\pi [G_t | S_t = s, A_t = a] \\
   & = E_\pi [ \sum^\infty_{k=0} \gamma^k R_{t+k+1} | S_t = s, A_t = a]
\end{align*} 
\section{Variational bayesian methods for deep learning}

\subsection{Practical Variational Inference for Neural Networks}
Since Variational Inference was first proposed \cite{hinton1993keeping} it was not used largely until \cite{graves2011practical} because of the complexity involved even for the simplest architectures such as radial basis[2] and single layer feedforward networks \cite{hinton1993keeping} \cite{barber1998radial} \cite{barber1998ensemble}. Graves reformulated Variational Inference as optimization of MDL (Minimum Description Length) Loss function Equation (12). \\\\
\cite{graves2011practical} derive the formula of $L^E(\beta,D)$, $L^C(\alpha,\beta)$ and their gradients for various choices of $q(\beta)$ and $P(\alpha)$. They limit themselves to diagonal posteriors of the form $q(\beta) = \Pi_{i=1}^{W} q_i(\beta_i)$ due to which the $L^C(\alpha,\beta) = \sum_{i=1}^{W} KL(q_i(\beta_i)||P(\alpha))$.

\subsubsection{Delta Posterior}
If the posterior distribution($q(\beta)$) is dirac delta which assigns 1 to a particular set of weights and 0 to any other weights $ L^E(\beta,D) = L^N(x,y,w)$ and $ L^C(\alpha,\beta) = L^C(\alpha,w) = -logP(w/\alpha)+C$. \\ 

This can be divided into three cases where prior is Uniform, Laplacian and Gaussian. 

\begin{itemize}
    \item \textbf{Uniform:-} If all realisable weights are equally likely then $ L^C(\alpha,\beta)$ is constant which makes it similar to maximum likelihood training as it effectively tries to optimize only $ L^E(\beta,D) $ loss.
    \item \textbf{Laplacian:-} In this case $\alpha=\{\mu,b\}$ and $P(w\alpha)=\prod_{i=1}^{W}\dfrac{1}{2b}exp\bigg(-\dfrac{|w_i-\mu|}{b}\bigg)$. Then $L^C(\alpha,w)$ is as follows
    \begin{align}
            L^C(\alpha,w)=Wln2b+\dfrac{1}{b}\sum_{i=1}^{W}|w_i-\mu|+C => \dfrac{\partial L^C(\alpha,w)}{\partial w_i}=\dfrac{sgn(w_i-\mu)}{b}
    \end{align}
    The optimal parameters for $\{\alpha,b\} = \{median(w), mean(|w-\mu|))\}$
    \item \textbf{Gaussian:-} In case of gaussian $\alpha=\{\mu,\sigma^2\}$ and $P(w\alpha)=\prod_{i=1}^{W}\dfrac{1}{\sqrt{2\pi\sigma^2}} exp\bigg(-\dfrac{(w_i-\mu)^2}{2\sigma^2}\bigg)$.
    Then $L^C(\alpha,w)$ is as follows
    \begin{align}
            L^C(\alpha,w)=Wln\sqrt{2\pi\sigma^2}+\dfrac{1}{2\sigma^2}\sum_{i=1}^{W}(w_i-\mu)^2+C => \dfrac{\partial L^C(\alpha,w)}{\partial w_i}=\dfrac{w_i-\mu}{\sigma^2}
    \end{align}
    The optimal parameters for $\{\alpha,\sigma^2\} = \{mean(w), mean((w-\mu)^2))\}$
\end{itemize}

\subsubsection{Gaussian Posterior}
If $\beta$ is diagonal gaussian distribution then each weight requires $\mu$ and $\sigma^2$. Neither $L^E(\beta,D)$ nor its derivatives can be computed exactly in this case. Hence \cite{graves2011practical} rely on sampling and use identities of gaussian expectations \cite{opper2009variational}:-
\begin{align}
    L^E(\beta,D) &\approx \dfrac{1}{S}\sum_{k=1}^{S} L^N(w^k,D) \\
    \dfrac{\partial L^E(\beta,D)}{\partial \mu_i} &= \bigg \langle \dfrac{\partial L^N(w,D)}{\partial w_i} \bigg \rangle_{w \sim q(\beta)} \approx \dfrac{1}{S} \sum_{k=1}^{S} \dfrac{\partial L^N(w^k,D)}{\partial w_i}\\
    \dfrac{\partial L^E(\beta,D)}{\partial \sigma_i^2} &= \dfrac{1}{2} \bigg \langle \dfrac{\partial^2 L^N(w,D)}{\partial w_i^2} \bigg \rangle_{w \sim q(\beta)} \approx \dfrac{1}{2S} \sum_{k=1}^{S} \bigg[ \dfrac{\partial L^N(w^k,D)}{\partial w_i}\bigg]^2
\end{align}
\begin{itemize}
    \item \textbf{Uniform:-} If the prior is uniform then minimizing $L(\alpha,\beta,D)$ is same optimizing for $L^N(w,D)$ with synaptic noise or weight noise.
    \item \textbf{Gaussian:-} If the prior is Gaussian then $\alpha=\{\mu,\sigma^2\}$ and 
    \begin{align*}
        L^C(\alpha,\beta) = \sum_{i=1}^{W}ln\dfrac{\sigma}{\sigma_i}+\dfrac{1}{2\sigma^2}\bigg[(\mu-\mu_i)^2+\sigma_i^2-\sigma^2\bigg]\\
        => \dfrac{\partial L^C(\alpha,\beta) }{\partial \mu_i} = \dfrac{\mu_i-\mu}{\sigma^2}, \dfrac{\partial L^C(\alpha,\beta) }{\partial \sigma_i^2}\dfrac{1}{2}\bigg[ \dfrac{1}{\sigma^2} - \dfrac{1}{\sigma_i^2} \bigg]
    \end{align*}
    The optimal parameters are:-
    \begin{align*}
        \hat{\mu} = \dfrac{\sum_{i=1}^{W}\mu_i}{W}; \hspace{1cm}
        \hat{\sigma^2} = \dfrac{\sum_{i=1}^{W}\bigg[\sigma_i^2 + (\mu_i-\hat{\mu})^2\bigg]}{W}
    \end{align*}
\end{itemize}

\subsection{Auto Encoding Variational Bayes}
\cite{kingma2013auto} show that reparameterization of the variational lower bound yields a simple differentiable unbiased estimator
of the lower bound. This SGVB (Stochastic Gradient Variational Bayes) is then straightforward to optimize using standard stochastic gradient ascent techniques. 
\subsubsection{Reparameterization Trick}
They express the random variable $z \sim q(z)$ (posterior) in terms of $x$ such that a deterministic mapping $z = g_\phi(\epsilon,x)$ where $\epsilon$ is a auxiliary variable with independent marginal $p(\epsilon)$ and $g_\phi(.)$ is some vector parameterized by $\phi$. As it is a deterministic mapping we know that:-

\begin{align*}
    \hspace{3cm} 
    q_\phi(z)\prod_i dz_i &= p(\epsilon) \prod_i  d\epsilon_i 
    \\\hspace{1cm}
    \int q(z)f(z)dz &= \int p(\epsilon)f(z)d\epsilon \\\hspace{1cm}
    \int q(z)f(z)dz &= \int p(\epsilon)f(g_\phi(\epsilon,x))d\epsilon 
    \\\hspace{1cm}
    \int q(z)f(z)dz &\approx \dfrac{1}{L}\sum_{l=1}^{L} p(\epsilon)f(g_\phi(\epsilon,x))d\epsilon^l \hspace{0.5cm}(where \hspace{0.3cm} \epsilon^l \sim p(\epsilon) ) \\
\end{align*}
For example in case of univariate Gaussian let $ z\sim \mathcal{N}(\mu,\sigma^2)$ and a valid reparameterization is $z=\mu+\sigma\epsilon$ where $\epsilon\sim \mathcal{N}(0,1)$. In this case they rewrite  $E_{\mathcal{N}(\mu,\sigma^2)}[f(z)] = E_{\mathcal{N}(0,1)}[f(\mu+\sigma\epsilon)] \approx
\dfrac{1}{L} \sum_{l=1}^{L} f(\mu+\sigma\epsilon^{l}).$ 

\subsubsection{SGVB Estimator and AEVB Algorithm}

The loss term can be written using the above reparameterization trick. Then resulting estimator is generic Stochastic Gradient Variational Bayes (SGVB) estimator. It is defined as follows
\begin{align*}
    L(\theta,\phi;x)&=E_{q(z/x)}[-logq(z/x)+logp(x,z)]\\
    L^A(\theta,\phi;x)&=\dfrac{1}{L}\sum_{l=1}^{L}[-logq(z^{(i,l)}/x^{i})+logp(x^{i},z^{(i,l)})] \approx L(\theta,\phi;x)
\end{align*} 
Here they consider that $\theta$ and $\phi$ are generative and vartaional parameters respectively.

Often KL term is integrable and in that case only $E_{q(z/x^i)}[logp(x^i/z)] $ requires sampling. This yields second version of SGVB estimator whose variance is less than generic one.  Then the loss term becomes as follows.
\begin{align*}
    L^B(\theta,\phi;x)&=-KL(q(z/x^i)||p(z))+\dfrac{1}{L}\sum_{l=1}^{L}[logp(x^{i}/z^{(i,l)})] \approx L(\theta,\phi;x)
\end{align*} 
They construct an estimator of the marginal likelihood lower bound of the full dataset, based on minibatches:
\begin{align*}
    L(\theta,\phi;X) \approx L^M(\theta,\phi;X^M) = \dfrac{N}{M} \sum_{i=1}^{M} (\theta,\phi;x^i)
\end{align*}
Here $X^M = \{x^i\}_{i=1}^{M}$ is a randomly drawn sample of M data points from full dataset containing N data points. The minibatch version of AEVB (Auto Encoding Variational Bayes) algorithm \cite{kingma2013auto} is in Algorithm \ref{alg:aevb}.

\begin{algorithm}[t]
\caption{Minibatch version of the Auto-Encoding VB (AEVB) algorithm.}
\renewcommand{\algorithmicforall}{\textbf{for each}}
\begin{algorithmic}
\State $\theta, \phi \gets$ Initialize parameters
\Repeat
\State $X^M \gets $ Random minibatch of $M$ datapoints (drawn from full dataset)
\State $\epsilon \gets $ Random samples from noise distribution $p(\epsilon)$
\State $g \gets \nabla_{\theta,\phi} L^{M}(\theta,\phi;X^M,\epsilon)$ (Gradients of minibatch estimator )
\State $\theta, \phi \gets $ Update parameters using gradients $g$ (e.g. SGD or Adagrad)
\Until {convergence of parameters $(\theta,\phi)$}
\\ \Return $\theta, \phi$
\end{algorithmic}
\label{alg:aevb}
\end{algorithm}

\subsection{Bayes by backprop}

\cite{blundell2015weight} Introduce a backpropagation compatible algorithm for learning a distribution of neural network weights known as Bayes by backprop.  Under certain conditions derivative of an expectation can be expressed as the expectation of derivative: \\\\
\textbf{Proposition 1.} \textit{Let $\epsilon$ be a random variable having a probability density given by q($\epsilon$) and let w = t($\theta$, $\epsilon$) where t($\theta$, $\epsilon$) is a deterministic function. Suppose further that the marginal probability density of w, q(w|$\theta$), is such that q($\epsilon$)d$\epsilon$ = q(w|$\theta$)dw. Then for a function f with derivatives in w:}
\begin{align*}
    \dfrac{\partial}{\partial \theta} E_{q(w/\theta)}[f(w,\theta)] = E_{q(\epsilon)}\bigg[ 
    \dfrac{\partial f(w,\theta)}{\partial w}\dfrac{\partial w}{\partial \theta} + \dfrac{\partial f(w,\theta)}{\partial \theta}
    \bigg]
\end{align*}
The above proposition is a generalization of Gaussian Reparameterization trick \cite{opper2009variational} \cite{kingma2013auto} \cite{rezende2014stochastic} . With the help of above proposition we can use monte carlo sampling to get a backpropogation like algorithm for variational bayesian inference of neural networks.

Bayes by backprop differs in many ways when compared to previous works.

\begin{itemize}
    \item They operate directly on weights while previous works operated on stochastic hidden units.
    \item There is no need for closed form of complexity cost for KL term.
    \item Performance is similar to using closed KL form.
\end{itemize}

\subsubsection{Gaussian Variational Posterior}
\cite{blundell2015weight} Illustrate bayes by backprop approach for gaussian variational posterior where the w is obtained by shifting by $\mu$ and scaling by $\sigma$. Further they paramterize $\sigma = log(1+exp(\rho))$. Hence the variational posterior parameters are $\theta=(\mu,\rho)$ and the posterior weights $w = t(\theta,\epsilon) = \mu + log(1+exp(\rho)) \bullet \epsilon$. The procedure is as follows:-

\begin{itemize}
    \item Sample $\epsilon \sim \mathcal{N}(0,I)$ 
    \item $w = t(\theta,\epsilon) = \mu + log(1+exp(\rho)) \bullet \epsilon$
    \item $\theta = (\mu,\rho)$
    \item Calculate loss $L(\theta,\phi;X)$
    \item Calculate gradients for $\theta$.
    \begin{align*}
        \bigtriangleup_{\mu} &=  \dfrac{\partial L(\theta,\phi;X)}{\partial w} + \dfrac{\partial L(\theta,\phi;X)}{\partial \mu}\\
        \bigtriangleup_{\rho} &=  \dfrac{\partial L(\theta,\phi;X)}{\partial w} \dfrac{\epsilon}{1+exp(-\rho)} + \dfrac{\partial L(\theta,\phi;X)}{\partial \rho}
    \end{align*}
    \item Update parameters:-
        \begin{align*}
            \mu \leftarrow \mu -\alpha\bigtriangleup_{\mu}\\
            \rho \leftarrow \rho -\alpha\bigtriangleup_{\rho}
        \end{align*}
\end{itemize}
Note that $\dfrac{\partial L(\theta,\phi;X)}{\partial \phi}$ is the exact gradient found by backprop algorithm of a neural network. They also  scale mixture prior of the form:-
\begin{align*}
    P(w) = \prod_j\pi\mathcal{N}(w_j/0,\sigma_1^2) + (1-\pi)\mathcal{N}(w_j/0,\sigma_1^2)
\end{align*}

\subsubsection{Mini batch KL re-weighting}
\cite{graves2011practical} proposed minibatch cost as
\begin{align*}
    L(\theta,\phi,X_i) = \dfrac{1}{M} KL(q(z/X_i)||P(w)) -  E_{q(z/X_i)}[logp(X_i/w)]
\end{align*}
Here $X_i$ is a minibatch from input($X$).\cite{blundell2015weight} observe that $\sum_iL(\theta,\phi,X_i)=L(\theta,\phi,X)$ and  weight the KL loss to likelihood cost in many ways, a generic way is as follows:-
\begin{align*}
    L(\theta,\phi,X_i) = \pi_i KL(q(z/X_i)||P(w)) -  E_{q(z/X_i)}[logp(X_i/w)]
\end{align*}
where $\pi \in [0,1]^M$ and $\sum_{i=1}^{M}\pi_i=1$. They found that $\pi_i = \dfrac{2^{M-i}}{2^{M}-1}$ weights the initial minibatches heavily influenced by KL Loss and later minibatches are heavily influenced by data to be more useful.

\subsection{Multiplicative Normalizing Flows}
Even though the lower bound optimization is straight forward for Bayes by Backprop \cite{blundell2015weight} the approximating capability is quite low, because it is similar to having a unimodal bump in high dimensional space. Even though there are methods such as \cite{gal2015bayesian} with mixtures of delta peaks and \cite{louizos2016structured} with matrix Gaussians that allow for nontrivial covariances among the weights, they are still limited.\\\\
Normalizing Flows (NF) have easy to compute Jacobian and they have been used to improve posterior approximation \cite{rezende2015variational} \cite{ranganath2016hierarchical}. Normalizing Flows(NF) can be applied directly to compute q(w) but they become quickly become really expensive and we will also lose the benefits of reparameterization.\\
\cite{louizos2017multiplicative} exploit well known "multiplicative noise" concept, e.g. as in Dropout . The approximate posterior can be parametrized with the following process:-

\begin{align*}
    z \sim q_{\phi}(z); \hspace{1cm} w \sim q_{\phi}(w/z)
\end{align*}
To allow for local reparameterization they parameterize the conditional distribution for the weights to be fully factorized Gaussian. Linear layers ( Algorithm \ref{alg:ff_mnf}) are assumed to be of this form :-

\begin{align*}
    q_{\phi}(w/z) = \prod_{i=1}^{D_{in}} \prod_{j=1}^{D_{out}} \mathcal{N}(z_i\mu_{ij},\sigma_{ij}^2)
\end{align*}
where $D_{in}$ and $D_{out}$ are input and output dimensions. For convolutional networks ( Algorithm \ref{alg:conv_mnf}) it is of the form:

\begin{align*}
    q_{\phi}(w/z) = \prod_{i=1}^{D_h} \prod_{j=1}^{D_w} \prod_{k=1}^{D_f} \mathcal{N}(z_k\mu_{ijk},\sigma_{ijk}^2)
\end{align*}
where Dh, Dw, Df are the height, width and number of
filters for each kernel. Here z is of very low dimension.

\begin{algorithm}[htb]
\caption{Forward propagation for each fully connected layer $h$. $M_w, \Sigma_w$ are the means and variances of each layer and $H$ is a minibatch of activations. Here $\odot$ is element wise multiplication.} 
\label{alg:ff_mnf}
\begin{algorithmic}[1]
\Require $H, M_w,\Sigma_w$
    \State $z_0 \sim q(z_0)$
    \State $z_{T_f} = NF(z_0)$
    \State $M_h = (H\odot z_{T_f})* M_w$
    \State $V_h = H^2\Sigma_w$
    \State $E \sim \mathcal{N}(0, 1)$
    \State return $M_h + \sqrt{V_h} \odot E$
\end{algorithmic}
\end{algorithm}

\begin{algorithm}[htb]
\caption{Forward propagation for each convolutional layer $h$. $N_f$ are the number of convolutional filters, $*$ is the convolution operator and we assume the [batch, height, width, feature maps] convention.} 
\label{alg:conv_mnf}
\begin{algorithmic}[1]
\Require $H, M_w, \Sigma_w$
    \State $z_0 \sim q(z_0)$
    \State $z_{T_f} = NF(z_0)$
    \State $M_h = H * (M_w \odot \text{reshape}(z_{T_f}, [1, 1, D_f]))$
    \State $V_h = H^2 * \Sigma_w$
    \State $E \sim \mathcal{N}(0, 1)$
    \State return $M_h + \sqrt{V_h} \odot E$
\end{algorithmic}
\end{algorithm}
They use masked RealNVP \cite{dinh2016density} with numerically stable updates introduced in Inverse Autoregressive Flow. 

\begin{align}
\*m \sim \text{Bern}(0.5); & \qquad \*h = \tanh(f(\*m \odot \*z_t))\nonumber\\
\!\mu = g(\*h); & \qquad \!\sigma = \sigma(k(\*h))\nonumber\\
\*z_{t+1} = \*m \odot \*z_t + & (1 - \*m) \odot (\*z_t \odot \!\sigma + (1 - \!\sigma) \odot \!\mu)\label{eq:nf_bnn}\\
\log \bigg|\frac{\partial\*z_{t+1}}{\partial\*z_t}\bigg| & = (1 - \*m)^T \log\!\sigma,\nonumber
\end{align}

Unfortunately parameterizing posterior in this form makes the lower bound intractable. But they make it tractable again by using auxiliary distribution $r(z/W)$ \cite{agakov2004auxiliary}. It is similar to doing variational inference in $p(D,w,z)$. The lower bound now becomes:-

\begin{align*}
    L(\theta,\phi;X) = E_{q(z,w)}[logp(y/x,w,z)+logp(w)+logr(z/w)-logq(w/z)-logq(z)]
\end{align*}
They parameterize $r(z/W)$ with inverse normalizing flows. In case of standard normal priors and fully factorized Gaussian posterior KL between prior and posterior can have a simpler closed form. For more details on $r(z/W)$ and closed form of KL, the reader can refer to \cite{louizos2017multiplicative}.

\subsection{Bayesian Hypernetworks}
A bayesian hypernetwork takes a random noise sample($\epsilon \sim \mathcal{N}(0,I)$) as an input and generates a approximate posterior $q(\theta)$using another primary network. The main idea in \cite{krueger2017bayesian} is to use an invertible hypernetwork  which helps  in estimating $-logq(\theta)$ in VI objective. \\\\
They use differentiable directed generative network (DDGN) as generative model for primary net parameters which are invertible. For being efficient in case of large primary networks they use normalization reparameterization :-

\begin{align*}
    \theta_j = gu, \hspace{0.4cm} u:=\dfrac{v}{||v||_2}, \hspace{0.4cm} g \in \mathbb{R}
\end{align*}
Here $\theta_j$ is the input weights for single unit j in primary network. They use the scaling factor $g$ from hypernets and $v$ from maximum likelihood estimate of v.

\section{Applications  in reinforcement learning}
 A fundamental problem in RL is the exploration/exploitation dilemma. This issue raises from the fact that the agent needs to maintain a balalce between exploring the environment and using the knowledge it acquired from this exploration. The two classic approaches to this task are $\epsilon $- greedy  \cite{sutton2018reinforcement} in which the agent takes a random action with some probability $\epsilon $ (a hyperparameter) and acts according to its learned policy with a probability $1 - \epsilon $ and entropy regularization \cite{williams1992simple} in which an entropy term is added to the loss which adds more cost to actions that dominate too quickly, favouring exploration. 
\subsection{Noisy Networks for exploration}
Noisy nets \cite{fortunato2017noisy} propose a simple and efficient way to tackle the above issue where learned perturbations of the
network weights are used to drive exploration. The method consists of adding gaussian noise to the last (Fully connected) layers of the network. The perturbations are sampled from a noise distribution. The variance of the perturbation is a parameter that is learned using gradients from the reinforcement learning loss function. \\ \\
Let $y = f_\theta (x)$ be a neural network which takes an input $x$ and outputs $y$. Noise parameters $\theta$ can be represented as $\theta = \mu + \Sigma \odot \epsilon$ where $\mu $ and $\Sigma$ are the learnable parameters, $\epsilon$ is a zero mean noise vector and $\odot$ denotes elementwise multiplication. Optimization occurs with respect to the parameters $(\mu , \Sigma)$. Consider a Fully connected layer of a neural network with $p$ inputs and $q$ outputs represented by 
\begin{align}
    y = wx + b
\end{align}
where $x \in \mathbb{R}^p$ is the input to this layer, $w \in \mathbb{R}^{q \times p}$ is the weight matrix, and $b \in \mathbb{R}^q$ is the bias. The noisy linear layer is defined as: 
\begin{align}
    y = (\mu ^ w + \sigma ^ w \odot \epsilon ^ w)x + \mu^b + \sigma ^ b \odot \epsilon ^ b
\end{align}
which is obtained by replacing $w$ and $b$ in Equation (?) by $\mu ^ w + \sigma ^ w \odot \epsilon ^ w$ and $\mu^b + \sigma ^ b \odot \epsilon ^ b$. \\\\
They apply this method to DQN and A3C algorithms, without using $\epsilon$-greedy or entropy regularization. They experiment with two ways to introduce noise into the model: 
\begin{itemize}
    \item Independent Gaussian Noise: every weight and bias of noisy layer is independent, where each entry $\epsilon^w_{i,j}$ if the random matrix $\epsilon^w$ is drawn from a unit Gaussian distribution.
    \item Factorized Gaussian Noise: By factorizing  $\epsilon^w_{i,j}$, $p$ unit Gaussian variable $\epsilon_i$ can be used for noise of the inptus and a $q$ unit Gaussian variable $\epsilon_j$ for nose of the output. 
\end{itemize}

\subsection{Deep Exploration via Bootstrapped DQN}
\cite{osband2016deep} presents another approach to replace the $\epsilon$-greedy exploration strategy, which focuses on the bootstrap approach to uncertainty for neural networks. The main idea is to encourage deep exploration by creating a new Deep Q - learning architecture that supports selecting actions from randomized Q-functions that are trained on bootstrapped data. 
\begin{figure}[ht]
    \centering
    \includegraphics[width=8cm]{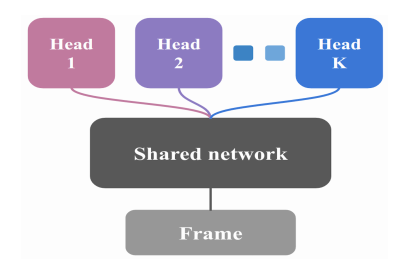}
    \caption{Shared network architecture.  \cite{osband2016deep}}
    \label{fig:dqn}
\end{figure} 
\\\\The network consists of a shared architecture with K bootstrapped heads that branch of independently, as shown in Figure \ref{fig:dqn}. Each head represents a Q function that is trained on a subset of the data.  Bootstrapped DQN modifies DQN \cite{mnih2015human} to approximate a distribution over Q-values via bootstrapping. At the beginning of each episode, a different Q-function is randomly sampled from a uniform distribution and it is used until the end of that episode.

\subsection{UCB Exploration via Q-Ensembles}
\cite{chen2017ucb} build on the Q-ensemble approach used by \cite{osband2016deep}. Instead of using posterior sampling for exploration, they use uncertainty estimates from Q-ensemble. Their method is based on Upper-confidence bounds (UCB) algorithms in bandit setting and constructs uncertainty estimates from Q-values. Q-function for policy $\pi$ is defined as $Q^\pi(s,a) = E_\pi [\sum^\infty_{t = 0}\gamma^t r_t | s_0 = s, a_0 = a]$. The optimal $Q^*$ function satisfies the Bellman equation
\begin{align*}
    Q^*(s,a) = E_{s' \sim T(.|s,a)}[r+\gamma . \mathop{max}_{a'}Q^*(s',a')]
\end{align*}
The ideal Bayesian approach to reinforcement learning is to maintain a posterior over the MDP. However, it is more tractable to maintain a posterior over the $Q^*$- function. Let the MDP be defined by the transition probability $T$ and reward function $R$. Assume that the agent samples $(s,a)$ according to a fixed distribution. The corresponding reward $r$ and next state $s'$ form a transition $\tau = (s,a,r,s')$ for updating the posterior of $(Q^*,T)$. Using Bayes' rule to expand the posterior, we get
\begin{align}
    \mathop{p}^{\sim} (Q^*,T|\tau) = \frac{p(Q^*,T).T(s'|s,a).p(s,a)}{Z(\tau)} \mathbbm{1} (Q^*,T)
\end{align}
where $\mathbbm{1}$ is the indicator function. The exact posterior update in the above equation is intractable due to the large space of $(Q^*,T)$. An extensive discussion on the several approximations made to the $Q^*$ posterior update can be found in \cite{chen2017ucb}. Using the outputs of K copies of independently initialized $Q^*$ functions, they construct a UCB by adding the empirical standard deviation $\mathop{\sigma}^{\sim}(s_t,a)$ of $\{Q_k(s_t,a)\}^K_{k=1}$ to the empirical mean of $\mathop{\mu}^{\sim}(s_t,a)$ $\{Q_k(s_t,a)\}^K_{k=1}$. The agent chooses the action that maximizes this UCB 
\begin{align*}
    a_t \in \mathop{argmax}_{a} \{\mathop{\mu}^{\sim}(s_t,a) + \lambda  \mathop{\sigma}^{\sim}(s_t,a) \}
\end{align*}
where $\lambda$ is a hyperparameter. They evaluate the algorithms on each Atari game of the Arcade Learning Environment. $Q$ ensemble approach outperforms Double DQN and bootstrapped DQN.

\section{Applications in Continual Learning}
Continual learning (also called life-long learning and incremental learning) is a very general form of online learning in which data continuously arrive in a possibly non i.i.d. way, tasks may change over time (e.g. new classes may be discovered), and entirely new tasks can emerge. In the continual learning setting, the goal is to learn the parameters of the model from a set of sequentially arriving datasets $\{x_t^{(n)}, y_t^{(n)} \}_{n=1}^{N_t}$. The posterior is defined by:
\begin{equation}
\label{eq:posterior}
p({\theta} | \mathcal{D}_{1:T}) \propto  p({\theta}) \prod_{t=1}^T \prod_{n_t=1}^{N_t} p(y_t^{(n_t)} | {\theta}, {x}_t^{(n_t)}) = p({\theta})  \prod_{t=1}^T  p(\mathcal{D}_t | {\theta})\propto p({\theta} | \mathcal{D}_{1:T-1}) p(\mathcal{D}_T | {\theta}).\nonumber
\end{equation}
Here $\mathcal{D}_t = \{ y_t^{(n)} \}_{n=1}^{N_t}$, $p(\theta)$ is the prior over $\theta$ and $T$ is the number of datasets.

\begin{equation}
q_t({\theta}) = \arg \min_{q \in \mathcal{Q}} \mathrm{KL} \Big( q({\theta}) ~\|~ \frac{1}{Z_t} q_{t-1}({\theta}) ~ p(\mathcal{D}_t | {\theta}) \Big), \text{ for } t = 1, 2, \ldots, T. \label{eqn:vcl}
\end{equation}
The loss in Equation \ref{eqn:vcl} is used in Varational Continuial Learning \cite{nguyen2017variational} where $q_0({\theta}) = p({\theta})$, t is the time step and $Z_t$ is the intractable normalizing constant of $q_{t-1}({\theta}) ~ p(\mathcal{D}_t | {\theta})$.\\ \\
In VCL the minimization at each step may be approximate and we may lose additional information so we also have a small representative set called as coreset for the previously observed tasks. The coreset can be selected by sampling random $K$ points from the $D_t$ dataset or we may try to select points which are spread in the input space. \\\\
They evaluate their model on Permuted MNIST, Split MNIST and Split notMNIST. For further details refer \cite{nguyen2017variational}

\section{Conclusion}
In this report, we review the major advances in Variational Bayesian methods from the perspectives of approaches and their applications in Reinforcement Learning and continual learning. Although this field has grown in the recent years, it remains an open question on how to make Variational Bayes more efficient.

\section{Acknowledgement}
The authors would like to thank Joan Bruna for his feedback and providing this opportunity.
\bibliographystyle{apalike}
\bibliography{references}

\end{document}